\pdfoutput=1
\documentclass[11pt]{article}

\usepackage[]{naacl2021}

\usepackage{times}
\usepackage{latexsym}
\usepackage{stfloats} 
\usepackage{graphicx} 
\usepackage{multirow}
\usepackage{algorithm}
\usepackage{ulem}
\usepackage{algpseudocode}
\usepackage{amsmath}
\usepackage{makecell} % Add this to your preamble
\usepackage{parskip}
\usepackage{listings}

\usepackage[T1]{fontenc}
\usepackage[utf8]{inputenc}

\usepackage{microtype}

\title{Hybrid AI for Responsive Multi-Turn Online Conversations with Novel Dynamic Routing and Feedback Adaptation}

\author{
 \textbf{Priyaranjan Pattnayak\textsuperscript{1}},
 \textbf{Amit Agarwal\textsuperscript{1}},
 \textbf{Hansa Meghwani\textsuperscript{2}},
 \\
 \textbf{Hitesh Laxmichand Patel\textsuperscript{1}},
 \textbf{Srikant Panda\textsuperscript{2}}
\\
\\
 \textsuperscript{1}OCI, Oracle America Inc.,
 \textsuperscript{2}OCI, Oracle India
\\
 \small{
   \textbf{Correspondence:} \href{mailto:priyaranjan.pattnayak@oracle.com}{priyaranjan.pattnayak@oracle.com}
 }
}

\begin{document}
% \nolinenumbers
\maketitle
\begin{abstract}
Retrieval-Augmented Generation (RAG) systems and large language model (LLM)-powered chatbots have significantly advanced conversational AI by combining generative capabilities with external knowledge retrieval. Despite their success, enterprise-scale deployments face critical challenges, including diverse user queries, high latency, hallucinations, and difficulty integrating frequently updated domain-specific knowledge. This paper introduces a novel hybrid framework that integrates RAG with intent-based canned responses, leveraging predefined high-confidence responses for efficiency while dynamically routing complex or ambiguous queries to the RAG pipeline. Our framework employs a dialogue context manager to ensure coherence in multi-turn interactions and incorporates a feedback loop to refine intents, dynamically adjust confidence thresholds, and expand response coverage over time. Experimental results demonstrate that the proposed framework achieves a balance of high accuracy (95\%) and low latency (180ms), outperforming RAG and intent-based systems across diverse query types, positioning it as a scalable and adaptive solution for enterprise conversational AI applications.

\end{abstract}

\section{Introduction}

Recent progress in NLP has drastically changed the landscape of conversational AI, and among such new state-of-the-art solutions, a class of Retrieval-Augmented Generation (RAG) systems has emerged. By combining large language models (LLMs) with separate information retrieval pipelines, RAG systems can generate contextually rich and factually grounded responses, which are necessary for knowledge-intensive applications \cite{lewis2020retrieval}. However, enterprise-scale conversational AI systems often face real-world challenges such as diverse user query patterns, varying levels of query complexity, and stringent latency requirements for seamless user experiences. High computational costs, susceptibility to hallucinations when retrieval is misaligned, and inefficiencies in managing frequently updated domain-specific knowledge further compound these challenges, particularly in dynamic environments like customer support \cite{guu2020retrieval, rocktaschel2020rethinking}. In practice, ensuring that such systems can scale while maintaining accuracy and low latency remains an industry pain point.

In contrast, classical intent-based chatbots are efficient in processing frequently asked questions (FAQ) and other predictable queries, thanks to using pre-defined responses. Their lightweight computational footprint and scalability also make them well-suited for high-confidence, domain-specific scenarios \cite{serban2017survey, shah2018building}. However, these systems are inherently rigid and often struggle with query diversity, especially when faced with ambiguous or context-dependent user interactions. In high-demand enterprise settings, the inability of intent-based systems to adapt quickly to evolving user needs or handle complex multi-turn dialogues \cite{shah2018building, zhao2020retrieval} results in inconsistent user experiences and increased operational overhead for manual updates. The inability to balance adaptability with efficiency underscores the need for hybrid systems that synergize the strengths of RAG and intent-based approaches.

In order to solve these challenges, we propose a novel hybrid framework that combines RAG systems with intent-based canned responses for dynamic, multi-turn customer service interactions. While prior works have explored combining RAG and intent-based systems independently, our contribution lies in a cohesive framework that not only integrates these elements but also introduces a dynamic confidence-based routing mechanism refined through user feedback. This mechanism ensures that query routing decisions are continuously optimized based on real-time user interactions, enabling a system that evolves and adapts without manual intervention. Additionally, our framework addresses scalability challenges by efficiently balancing computational resources, making it particularly suited for enterprise-scale applications where latency and accuracy are paramount.
Our approach utilizes a dynamic query routing mechanism that evaluates the intent confidence level of user queries:
\begin{itemize}
    \item \textit{High-confidence queries} are resolved using predefined canned responses to ensure low latency and computational efficiency.
    \item \textit{Low-confidence or ambiguous queries} are routed to the RAG pipeline, enabling contextually enriched responses generated from external knowledge.
\end{itemize}

The framework is further enhanced with a dialogue context manager, keeping track and managing evolving intents across multiple turns, ensuring consistent and coherent interactions. Additionally, a feedback loop continuously refines the intent repository, adapting to emerging user needs and expanding response coverage over time. Our system is designed to meet enterprise latency standards, delivering responses within an acceptable threshold (sub-200ms latency and high turn efficiency), thereby ensuring user engagement and satisfaction in real-time applications\cite{pattnayak2024survey}.

\paragraph{Our Contributions}
This work makes the following key contributions:
\begin{enumerate}
    \item \textbf{Hybrid Conversational Framework:} We propose a novel architecture which combines RAG systems with intent-based canned responses; the queries are routed dynamically for optimizing response latency and computational cost without compromising accuracy.
    \item \textbf{Multi-Turn Dialogue Management:} We introduce a dialogue context manager which can track the evolving user intents and guarantee coherence in responses over multiple turns, thus addressing a key gap in the current systems.
    \item \textbf{Feedback-Driven Adaptability:} Our framework incorporates a feedback loop to enable continuous refinement of intents, canned responses and confidence thresholds, thereby improving system adaptability and coverage for real-world applications.
    \item \textbf{Comprehensive Evaluation:} Extensive experiments on synthetic and real-world datasets demonstrate significant improvements in accuracy, latency, and cost efficiency compared to state-of-the-art baselines.
    \item \textbf{Real-World Applicability:} Our framework is designed for enterprise-scale deployment, handling diverse user queries efficiently, from repetitive FAQs to complex knowledge-based questions, while adhering to industry latency standards critical for user retention.
\end{enumerate}

By addressing key challenges faced by enterprise conversational AI systems, such as query diversity, dynamic knowledge updates, and real-time latency requirements, our proposed framework offers a scalable, adaptive, and efficient solution. This work advances task-oriented dialogue systems, particularly in domains where multi-turn interactions and dynamic knowledge management are essential for operational success.

\section{Related Work}

\subsection{Retrieval-Augmented Generation (RAG)}
Recent advancements in RAG have enhanced contextual retrieval and generative capabilities, improving incident resolution in IT support ~\cite{isaza2024retrievalaugmentedgenerationbasedincident}, question-answering systems, and domain-specific chatbots~\cite{veturi2024ragbasedquestionansweringcontextual}. Research on noise handling ~\cite{Cuconasu_2024,meghwani2025hardnegativeminingdomainspecific} and reinforcement learning~\cite{kulkarni2024reinforcementlearningoptimizingrag} further optimizes RAG for precision and adaptability in complex applications. By retrieving relevant documents during inference, RAG systems mitigate common LLM challenges such as hallucinations and outdated knowledge \cite{lewis2020retrieval, guu2020retrieval}. These systems are particularly effective for knowledge-intensive tasks where accuracy and factual grounding are critical\cite{pattnayak9339review}.

Despite their effectiveness, RAG systems face significant challenges, including high computational costs and latency due to the dual retrieval and generation processes. Enterprise settings also pose unique challenges, such as diverse user queries, latency constraints, and evolving domain knowledge needs~\cite{lewis2020retrieval, pattnayak2025improvingclinicalquestionanswering}. Moreover, most existing RAG systems are optimized for single-turn interactions and struggle with maintaining coherence in multi-turn dialogues, where evolving user intents require dynamic retrieval and contextual adaptation \cite{rocktaschel2020rethinking}. Recent efforts to optimize RAG pipelines, such as multi-stage retrieval systems \cite{lee2020latency} and model distillation \cite{guu2020retrieval}, have reduced latency but do not address the complexities of multi-turn interactions \cite{sanh2020feedback}.

\subsection{Intent-Based Chatbots}
Intent-based chatbots work well for predictable \cite{pattnayak2017predicting}, high-confidence queries by mapping user inputs to predefined intents. These systems are widely used in domains like customer support, where they efficiently handle FAQs and repetitive queries with minimal computational overhead \cite{serban2017survey, shah2018building}. However, their reliance on predefined intents limits their adaptability to ambiguous or evolving queries\cite{pattnayak2025tokenizationmattersimprovingzeroshot}, particularly in multi-turn conversations \cite{michelson2020advances, friedrich2020context}.

Recent developments have involved the inclusion of transformer-based models to enhance intent recognition and increase coverage \cite{michelson2020advances}. However, these methods are resource-heavy, as they require a lot of labeled data and computational resources, which makes scalability quite limited for dynamic domains. 
\begin{table*}[t!]
\centering
\renewcommand{\arraystretch}{1.5} % Increase row height
\resizebox{\textwidth}{!}{%
\begin{tabular}{|p{3.5cm}|p{4cm}|p{4cm}|p{3.5cm}|p{3.5cm}|}
\hline
\textbf{Approach}          & \textbf{Strengths}                       & \textbf{Weaknesses}                    & \textbf{Multi-Turn Support} & \textbf{Feedback Adaptation} \\ \hline
RAG Systems                & Accurate, dynamic responses              & High latency, computationally expensive & Limited                     & No                            \\ \hline
Intent-Based Chatbots      & Efficient, low latency                   & Rigid, poor adaptability               & No                          & No                            \\ \hline
Hybrid RAG-Intent Systems  & Balance between efficiency and flexibility & Limited multi-turn and feedback mechanisms & Partial                     & No                            \\ \hline
Proposed Framework         & Low latency, multi-turn adaptable        & Scalability challenges                  & Yes                         & Yes                           \\ \hline
\end{tabular}%
}
\caption{Comparison of Existing Approaches and the Proposed Framework.}
\label{tab:comparison}
\vspace{-1em}
\end{table*}
\subsection{Hybrid Approaches}
Hybrid retrieval systems integrating lexical search (e.g., BM25 ~\cite{Robertson1994}) and semantic search (e.g., dense embeddings via FAISS~\cite{douze2024faisslibrary}) effectively balance speed and semantic depth~\cite{agarwaletal2025fs,pattnayak2025clinicalqa20multitask}, improving retrieval accuracy ~\cite{mitra2021hybrid, hernandez2020efficient}.

In conversational AI, hybrid approaches integrating RAG with intent-based responses have emerged to address limitations in single-mode systems by enhancing flexibility and efficiency ~\cite{ bordes2020context}. Prior works, such as \cite{gao2020search, zhao2020retrieval, patel2024llm}, have explored blending retrieval-augmented pipelines with canned responses to improve response efficiency and contextual depth. However, these systems are primarily designed for single-turn interactions and do not address the complexities of multi-turn dialogues, where query context evolves dynamically \cite{pattnayak2025improvingclinicalquestionanswering}. While existing research relies on static threshold-based routing, the integration of adaptive threshold driven routing and response generation for real-time, multi-turn applications remains an under explored area with significant potential for optimization.

\subsection{Positioning of This Work}
While prior research has advanced RAG systems, intent-based chatbots, and hybrid architectures, key limitations remain. RAG systems excel in generating contextually rich responses but struggle with coherence in multi-turn conversations, high latency, and computational costs \cite{lewis2020retrieval, rocktaschel2020rethinking}. Intent-based chatbots are efficient but lack flexibility for ambiguous or evolving queries in dynamic settings \cite{serban2017survey, agarwal2024enhancing}. Hybrid systems balance efficiency and adaptability but often fail to track dialogue context or refine responses dynamically based on user feedback \cite{gao2020hybrid}. Table \ref{tab:comparison} summarizes the key differences between the existing work and our proposed framework.

This work addresses real-world challenges by proposing a hybrid framework that integrates RAG systems with intent-based canned responses. It uses dynamic query routing to handle high-confidence queries efficiently with canned responses while relying on RAG pipelines for complex cases. A dialogue context manager ensures coherence in multi-turn interactions, and a real-time feedback loop enables continuous refinement of intents, thresholds and canned responses. For instance, in an enterprise customer support setting, our system efficiently handles high-frequency queries such as, \textit{“How do I reset my password?”} using canned responses with minimal latency (under 200ms), ensuring quick resolution for routine tasks. In contrast, more complex queries such as, \textit{“Can you help me troubleshoot a payment gateway integration issue with API X?”} are dynamically routed to the RAG pipeline, leveraging external documentation and past incident reports to generate accurate responses. This adaptability is further evident when users provide feedback on response quality, prompting the system to refine its intent classification and adjust confidence thresholds for future queries. Unlike existing systems that either focus on single-turn interactions or static routing and struggle with multi-turn dialogue management, our framework continuously adapts to diverse queries and user needs, optimizing latency and scalability.

By focusing on these critical aspects, this framework advances the state-of-the-art in task-oriented dialogue systems, particularly for enterprise-scale applications where efficiency, scalability, and adaptability are paramount.
\begin{figure*}[t!]
    \centering
    \includegraphics[width=\textwidth]{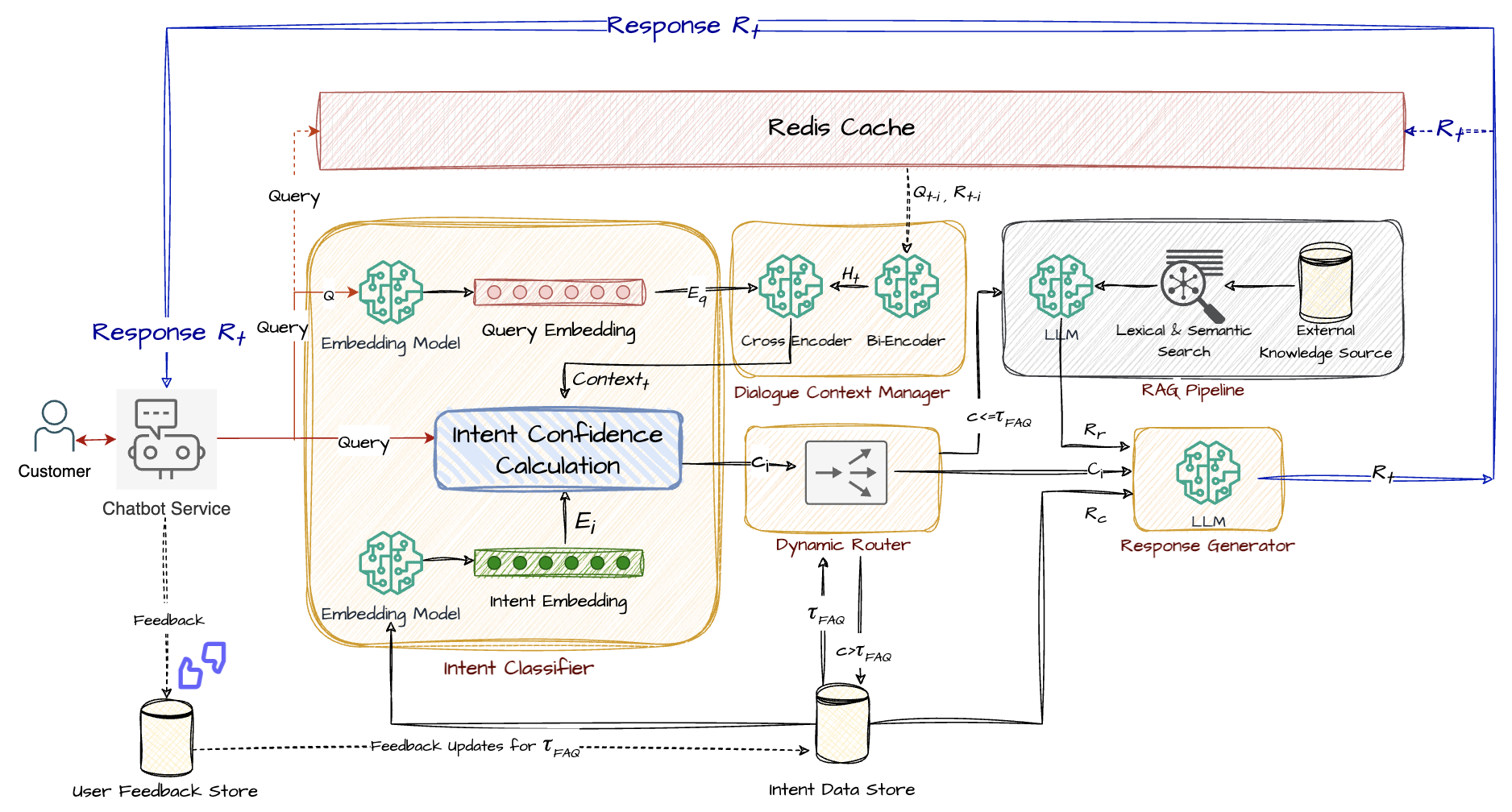} % Placeholder for system architecture image
    \caption{High-level Architecture of the Hybrid Framework.}
    \label{fig:system_architecture}
    \vspace{-1em}
\end{figure*}

\section{Proposed Framework}

The proposed framework integrates the efficiency of intent-based canned responses with the contextual richness and adaptability of Retrieval-Augmented Generation (RAG) systems~\cite{lewis2020retrieval, gao2020modular,patel2025sweeval}. By dynamically routing queries based on intent confidence and leveraging user feedback for adaptive refinement, the framework addresses latency, accuracy, and scalability challenges while maintaining coherence across multi-turn interactions. Figure~\ref{fig:system_architecture} illustrates the architecture with key modules, data flow and a Redis Cache which stores frequently accessed intents and responses for faster retrievals.

\begin{algorithm*}
\caption{Context-Aware Intent Confidence Calculation}
\label{alg:context_aware_intent}
\begin{algorithmic}[1]
\Require Query $Q$, Set of Intent Embeddings $\{E_1, E_2, \dots, E_n\}$, Historical Context Embeddings $\text{H}_{t}$
\Ensure Highest Confidence Score $c$, Corresponding Intent: $\text{Intent}_{\text{max}}$

\State \textbf{Step 1: Calculate Query Embedding}
\State $\text{E}_q \gets \text{BERT}(Q)$
\State \textbf{Step 2: Contextual Query Embedding}
\State $\text{Context}_t \gets \phi( E_q, \text{H}_{t})$
    \Comment{Augment query embedding with historical context}

\State \textbf{Step 3: Confidence Calculation}
\For{each intent embedding $E_i$ in $\{E_1, E_2, \dots, E_n\}$}
    \State $c_i \gets \text{CosineSimilarity}(\text{Context}_t, E_i)$
    \Comment{Compute similarity for intent $i$}
\EndFor

\State \textbf{Step 4: Find Best Match}
\State $c \gets \max(c_i)$
    \Comment{Highest confidence score}
\State $\text{Intent}_{\text{max}} \gets \text{argmax}_i(c_i)$
    \Comment{Intent corresponding to $c$}

\State \textbf{Output:} $c$, $\text{Intent}_{\text{max}}$
\end{algorithmic}
\end{algorithm*}

\subsection{Key Modules}

The framework comprises the following key components, each designed to address specific challenges in multi-turn dialogue systems:

\paragraph{Intent Classifier.} The Intent Classifier utilizes a fine-tuned BERT model~\cite{devlin2019bert} to encode user queries into semantic embeddings extracted from last layer of the model. See Appendix \ref{sec:in-house-data} for datatset detail. Confidence scores ($c$) are calculated by comparing the query embedding with predefined intent embeddings:
% \[
% c_i = \text{CosineSimilarity}(\text{Embedding}_q^{\text{context}}, E_i), \quad c = \max(c_i), \quad \text{Intent}_{\text{max}} = \text{argmax}_i(c_i).
% \]
Based on $c$, the query is classified as:
\begin{itemize}
    \item $c > 0.85$: \textbf{FAQ (Canned Response)}.
    \item $0.5 < c \leq 0.85$: \textbf{Contextual}.
    \item $c \leq 0.5$: \textbf{Out-of-Domain}.
\end{itemize}

The above thresholds are default for the system which are updated based on the user-feedback on the fly. Algorithm~\ref{alg:context_aware_intent} provides the pseudo-code for the classification process, which incorporates historical context from the Dialogue Context Manager.

\paragraph{Dialogue Context Manager.} The module tracks dialogue history using embeddings of prior queries and responses, stored in a sliding window. For multi-turn interactions, historical context embeddings are computed dynamically:
\[
\text{H}_t = \psi\big(\{(Q_{t-i}, R_{t-i}) \mid i = 1, .. n\}\big)
\]
where \(\psi\) represents a bi-encoder (in-house architecture) that computes the embeddings by appending prior context, queries, and responses into a string. \(Q_{t-i}\) and \(R_{t-i}\) represents previous query and corresponding responses within a chat session. The aggregated historical context \(\text{H}_t\) is then used to compute the current contextual query embedding:
\[
\text{Context}_t = \phi(E_q, \text{H}_t)
\]
Here, \(\phi\) represents a lightweight transformer block (in house cross-encoder) to compute attention, \(E_q\) is the current query embedding. Relevant historical context embedding is appended to the current query embedding for downstream processing.

\paragraph{Dynamic Routing.} The module checks the confidence ($c$) of the classified intent: $\text{Intent}_{\text{max}}$, against the threshold ($\tau_{FAQ}$) of the particular intent in the Intent Data Store. $\tau_{FAQ}$ for each intent is dynamically updated with user-feedback.

\paragraph{Response Generator.} The module refines the final response to user by either blending the static canned responses ($R_c$) with dynamic RAG outputs ($R_r$) using a language module or directly passing the $R_c$ or $R_r$ to the user based on the Dynamic Router. 

\paragraph{Feedback Mechanism.} Explicit (ratings) and implicit (e.g., query refinements) feedback is logged and used to refine thresholds, intents, and response mappings. New intents are created for recurring unhandled queries. Specifically, recurring unhandled queries are logged and grouped based on semantic similarity. When a threshold number of similar unresolved queries is reached in a group, the system automatically flags for creation of a new intent and response. Explicit user feedback is collected via a post-response prompt in the chat interface, allowing users to rate responses positive or negative (thumbs up or thumbs down), which dynamically updates the system's confidence thresholds every 100 interactions.

\subsection{Workflow}

The framework integrates query classification, response routing, multi-turn handling, and feedback adaptation into a cohesive workflow:

% \paragraph{Dynamic Routing.} This module determines the processing pathway for each query based on the confidence score ($c$) of the classified intent $\text{Intent}_{\text{max}}$ from the Intent Classifier. The threshold $\tau_{\text{FAQ}}$ for each intent, stored in the Intent Data Store, is dynamically updated using user feedback. The routing logic is as follows:

\paragraph{Query Classification.} Queries are classified into types (FAQ, Contextual, or Out-of-Domain) based on the confidence score $c$ from the Intent Classifier and the threshold $\tau_{\text{FAQ}}$ \& $\tau_{\text{Out-of-Domain}}$ for each intent, stored in the Intent Data Store, which is dynamically updated with the user feedback. The classification logic is as follows:
\begin{itemize}
    % \item \textit{Canned Response (FAQ):} If $c > \tau_{\text{FAQ}}$, the query is resolved using a predefined canned response for rapid resolution.
    % \item \textit{RAG Response (Out-of-Domain):} If $c \leq \tau_{\text{Out-of-Domain}}$, the query is routed exclusively to the RAG pipeline for domain-specific response generation.
    % \item \textit{Hybrid Response (Contextual):} If $\tau_{\text{Out-of-Domain}} < c \leq \tau_{\text{FAQ}}$, the query is processed by both canned responses and the RAG pipeline. The Response Generator then combines the outputs.
    \item \textit{FAQ:} If $c > \tau_{\text{FAQ}}$, the query is resolved using a predefined canned response for the intent.
    \item \textit{Out-of-Domain:} If $c \leq \tau_{\text{Out-of-Domain}}$, the query is routed exclusively to the RAG pipeline for domain-specific response generation.
    \item \textit{Contextual:} If $\tau_{\text{Out-of-Domain}} < c \leq \tau_{\text{FAQ}}$, the query is processed by both canned responses for the intent and the RAG pipeline. The Response Generator then combines the outputs.
\end{itemize}

\paragraph{Response Routing.} The final response for the user is based on the query classification. The response generation varies by query type:
\begin{enumerate}
    \item \textit{Canned Response (FAQ):} The predefined the canned response for the intent is passed directly to the user  for rapid resolution.
    \item \textit{RAG Response (Out-of-Domain):} The RAG output is passed as is, ensuring the most contextually rich response for undefined intents.
    \item \textit{Hybrid Response (Contextual):} Both the canned response and the RAG output are retrieved and combined into a unified response using a language model (LLM):
    \[
    R_f = \text{LLM}(c \cdot R_c, (1-c) \cdot R_r),
    \]
    where $c$ is the confidence of the $\text{Intent}_{\text{max}}$, passed to the LLM in to the prompt to ensures coherence and contextual alignment in the final response.
\end{enumerate}

\paragraph{Multi-Turn Handling.} Context tracking ensures coherence in multi-turn interactions by retrieving and appending the most relevant embeddings from dialogue history.

\paragraph{Feedback-Driven Adaptability.} User feedback dynamically influences system thresholds and intent mappings. The threshold for FAQs ($\tau_{\text{FAQ}}$) is adjusted based on feedback trends, ensuring that frequently misclassified queries are handled appropriately. The update mechanism follows::
\[
\tau_{\text{FAQ}} = \tau_{\text{FAQ}} + \lambda \cdot (\text{NFR} - \text{PFR}),
\]
where:
\begin{itemize}
    \item \textbf{NFR}: Negative Feedback Rate.
    \item \textbf{PFR}: Positive Feedback Rate.
    \item $\lambda$: Scaling factor controlling the sensitivity of the adjustment.
    \item $\tau_{\text{FAQ}}$: By default is set to 0.85 whenever the intents (and dependent intents) are updated in intent data store.
\end{itemize}

High negative feedback increases the threshold, reducing the likelihood of misclassification as FAQs, while positive feedback reduces the threshold to favor FAQ classification. Threshold for Out-of-Domain queries ($\tau_{\text{Out-of-Domain}}$) is kept constant at 0.5 to prevent over-restricting or over-generalizing OOD classification. This adaptive threshold mechanism ensures that the system remains responsive to user feedback while maintaining stability in query classification.  Further details are provided in Appendix \ref{sec:extended_workflow}

\subsection{Prototype Implementation}
The framework is implemented as a modular system using microservices:
\begin{itemize}
    \item \textbf{Frontend:} Built with React.js for user interaction and feedback collection.
    \item \textbf{Backend:} Flask microservices handle query classification, retrieval, and feedback processing~\cite{grinberg2018flask}.
    \item \textbf{Storage:} OCI (Oracle Cloud Infrastructure) Opensearch stores canned responses \& external knowledge base, while FAISS and dense embeddings support retrieval~\cite{karpukhin2020dense}. 
    \item \textbf{Memory Cache:} A memory-augmented module maintains embeddings of prior inter-actions in OCI Cache (Managed Redis), allowing the system to retain relevant historical context across multiple dialogue turns.
    \item \textbf{Model Deployment:} Models (e.g., BERT, Encoder, Cross-Encoder, GPT-3 \& other proprietary LLMs) are deployed using in-house architecture and OCI Gen AI Service for scalability.
\end{itemize}

\section{Experiment and Results}
The hybrid framework was evaluated on four metrics: accuracy, response latency, cost efficiency, and turn efficiency. These evaluations spanned in-house datasets of live customer queries, and scalability tests. Table~\ref{tab:results} summarizes overall results, while Table~\ref{tab:framework_comparison} in the appendix provides category-wise performance.

\subsection{Experimental Setup}
The evaluation dataset comprised 10,000 queries, categorized as :- 
a) \textit{Predefined FAQ Queries (40\%)}: High-confidence queries resolved via canned responses, b) \textit{Contextual Queries (30\%)}: Queries requiring both canned \& RAG responses, and c) \textit{Out-of-Domain Queries (30\%)}: Undefined intents handled exclusively by RAG pipeline.

For multi-turn interactions, 20\% of queries included follow-ups designed to assess context retention. Scalability tests evaluated performance with dataset sizes up to 50,000 queries, preserving category proportions. Results are shown in Table \ref{tab:scalability}.

\paragraph{Evaluation Metrics}
The system was assessed using the following metrics:
\begin{itemize}
    \item \textbf{Accuracy}: Percentage of correctly resolved queries across predefined FAQs, contextual queries, and out-of-domain scenarios. Accuracy is a fundamental evaluation metric in retrieval-based and generative NLP models~\cite{karpukhin2020dense, lewis2020retrieval}, ensuring that responses align with the intended knowledge base. We determine accuracy using a cosine similarity metric, as used in prior works on retrieval-based QA systems~\cite{reimers2019sentencebert}. For \textit{Predefined FAQ}, the framework has to fetch the correct FAQ, leading to a 100\% cosine similarity. For \textit{Contextual} and \textit{Out-of-Domain Queries}, the generated resposne needs to be similar (90\%) to annotated ground truth answer.
    
    \item \textbf{Response Latency}: Average response time in milliseconds taken to generate responses. Response latency is crucial in real-time conversational AI applications, as delays directly impact user experience~\cite{shuster2021retrievalaugmented,agarwal2025mvtamperbenchevaluatingrobustnessvisionlanguage}. Faster response times enhance engagement, making this metric essential for evaluating system efficiency.
    \item \textbf{Cost Efficiency (CE)}: A normalized measure of cost efficiency, defined as:
    
{\small
\[
\text{CE} = 
\min\left(
    1, 
    \frac{\text{Latency}_{\text{baseline}}}{\text{Latency}_{\text{proposed}}} 
    \times 
    \frac{\text{Accuracy}_{\text{proposed}}}{\text{Accuracy}_{\text{baseline}}}
\right)
\]
}

Inspired by cost-aware NLP evaluations~\cite{tay2023efficientllms}, this metric balances accuracy and latency trade-offs. It ensures that the proposed framework maintains or improves accuracy while reducing computational costs, a key factor in large-scale AI deployment.

    % \item \textbf{Context Retention}: The percentage of follow-up queries with preserved dialogue context.
    \item \textbf{Turn Efficiency}: Average number of turns required to resolve a query in a conversation:
    \[
    \text{Turn Efficiency} = \frac{\text{Total Turns}}{\text{Resolved Queries}}
    \]
    Turn efficiency measures conversational conciseness, ensuring that the system minimizes unnecessary back-and-forth interactions~\cite{serban2017survey}. A lower number of turns per resolved query indicates a more efficient dialogue system, reducing user dissatisfaction and operational overhead.
    % Conversations which are not resolved, are not taken into accoun 
\end{itemize}

\begin{table*}[t!]
\centering
\renewcommand{\arraystretch}{1.5}
\scalebox{0.8}{
\begin{tabular}{|p{4.5cm}|c|c|c|c|}
\hline
\textbf{Framework}              & \textbf{Accuracy (\%) ($\uparrow$)} & \textbf{Response Latency (ms) ($\downarrow$)} & \textbf{Cost Efficiency ($\uparrow$)} & \textbf{Turn Efficiency ($\downarrow$)} \\ \hline
Canned-Response (Baseline)         & 53                    & \textbf{68}                   & \textbf{1.0}             & \textbf{NA}                       \\ \hline
RAG Pipeline                    & \underline{91}           & 380                           & 0.3                      & 2.3                       \\ \hline
Proposed Framework       & \textbf{95}           & \underline{180}                           & \underline{0.7}             & \underline{1.7}             \\ \hline
\end{tabular}
}
\caption{Evaluation Results for Canned-Response (Intent) Systems, RAG, and Proposed Frameworks. Metrics represent averages across the evaluation dataset. The desired direction for improvement: ($\uparrow$) higher is better, ($\downarrow$) lower is better. Turn Efficiency is not available for Canned-Response as it lacks multi-turn capabilities.}
\label{tab:results}
\vspace{-1em}
\end{table*}

\subsection{Results and Analysis}
\paragraph{Overall Performance.} Table~\ref{tab:results} compares the proposed framework with baseline systems. Our proposed framework achieves a balance of high accuracy (95\%) and low latency (180ms), outperforming the canned-response system and the RAG pipeline's accuracy. The proposed system reduces the chances of hallucination for the most common user queries by leveraging canned responses hence outperforming accuracy of just RAG pipeline's. 

\paragraph{Category-Specific Insights.} Table~\ref{tab:framework_comparison} (Appendix \ref{appendix:detail_comparison}) highlights performance variations across query types:
\begin{itemize}
    \item \textbf{FAQs}: Similar accuracy compared to the canned-response system, with a ~82\% reduction in latency compared to RAG Pipeline.
    \item \textbf{Contextual Queries}: Accuracy improved over 47\% compared to canned-response system, with over 50\% reduction in latency compared to RAG Pipeline with similar accuracy.
    % The proposed framework achieved 96\% accuracy, blending canned responses and contextual depth.
    \item \textbf{Out-of-Domain Queries}: The RAG pipeline and our proposed framework exceed the baseline intent-based system's accuracy by over 85\%, as intent systems default to fallback responses for out-of-domain queries.
    % While RAG led with 75\% accuracy, the hybrid system followed closely at 78\%, with significantly reduced latency.
\end{itemize}

\paragraph{Scalability.}  The hybrid framework's scalability was evaluated under query loads ranging from 1,000 to 50,000. We observed graceful performance degradation under increasing query loads. Accuracy remains within enterprise-grade thresholds (92\% at 50,000 queries), while latency increases proportionally due to retrieval bottlenecks. Table~\ref{tab:scalability} summarizes the results, demonstrating the frameworks ability to maintain balanced performance in terms of accuracy, latency, and cost efficiency under increasing concurrent loads.

\paragraph{Cost Efficiency.} Proposed framework demonstrates effective trade-offs, achieving a CE score of 0.7 compared to 0.3 for RAG. The introduction of dynamic query routing minimizes computational overhead for high-confidence queries.

% \paragraph{Context Retention and Turn Efficiency.} High context retention (90\%) and turn efficiency (1.7) highlight the framework's ability to maintain coherence and minimize dialogue complexity, outperforming both baselines.

\paragraph{Turn Efficiency.} Turn efficiency (1.7) highlight the framework's ability to maintain coherence and minimize dialogue complexity while trying to resolve queries, relatively outperforming both baselines when compared in conjunction with accuracy and response latency.

\paragraph{Multi-Turn Interaction Analysis}
With 20\% (2,000) queries including follow-up interactions, the dialogue context manager maintained high coherence in these multi-turn interactions, effectively tracking evolving user intents and ensuring context continuity. Minor context drift was observed in sessions exceeding 10 turns, indicating that optimizing context management for prolonged dialogues remains an area for future improvement. See Appendix \ref{appendix:Failure_Cases} for common failure scenarios and error analysis.

% \section{Additional Evaluation Details}
% \label{sec:evaluation_appendix}
% For the evaluation of multi-turn adaptation and hybrid routing, we conducted the following:
% \begin{itemize}
%     \item \textbf{Synthetic Query Dataset:} Generated 20,000 multi-turn conversations spanning predefined FAQs, contextual queries, and out-of-domain scenarios.
%     \item \textbf{Turn-Level Metrics:} Evaluated context retention and turn efficiency by measuring the accuracy of responses across consecutive turns.
%     \item \textbf{Scalability Testing:} Simulated query loads ranging from 1,000 to 50,000 queries to measure system latency and throughput.
% \end{itemize}

\begin{table}[h!]
\centering

\renewcommand{\arraystretch}{1.5}
\scalebox{0.7}{
\begin{tabular}{|l|c|c|c|}
\hline
\textbf{Query Load} & \textbf{Accuracy (\%)} & \textbf{Latency (ms)} & \textbf{Cost Efficiency} \\ \hline
1,000                         & 96                    & 174                   & 0.77                      \\ \hline
5,000                         & 96                    & 177                   & 0.74                      \\ \hline
10,000                        & 95                    & 180                   & 0.71                      \\ \hline
20,000                        & 94                    & 186                   & 0.70                      \\ \hline
50,000                        & 92                    & 193                   & 0.69                      \\ \hline
\end{tabular}}
\caption{Scalability Results for the proposed Framework. Query Load indicates the number of queries processed in the evaluation.}
\label{tab:scalability}
\end{table}

\subsection{Error Analysis:}  
We conducted a manual error analysis on 500 dialogue samples covering diverse user intents. Only 32 (6\%) samples were found erroneous. Three independent annotators with subject matter expertise in Oracle cloud customer support evaluated these dialogue samples with an inter-annotator agreement of 0.91. Errors were categorized into four main types: 1) Edge Cases in Intent Classification (21\%) due to subtle semantic differences, 2) Long Multi-Turn Dialogues (35\%) where latency and context tracking issues arose, 3) Retrieval Inaccuracy (25\%) from incomplete or outdated document retrieval, and 4) Feedback Misalignment (19\%) due to misinterpretation of user feedback. Future work to remediate these could include refining fallback strategies, optimizing context management, regular index updates, and context-aware feedback processing. Further details are provided in Appendix~\ref{appendix:Failure_Cases}.

\subsection{Final Insights and Implications}
The evaluation metrics, error analysis and scalability underscore the proposed framework's effectiveness:
\begin{itemize}
    \item \textbf{Efficiency-Accuracy Trade-offs}: Dynamic query routing achieves optimal balance between computational cost and response quality.
    \item \textbf{Multi-Turn Adaptability}: Superior context retention validates its applicability in complex dialogue scenarios.
    \item \textbf{Scalability and Robustness}: Modular design ensures operational resilience under high query loads.
\end{itemize}

\section{Conclusion}
We proposed a hybrid conversational framework that integrates intent-based canned responses with Retrieval-Augmented Generation (RAG) systems, explicitly designed to handle multi-turn interactions. The framework dynamically routes queries based on intent confidence, ensuring low latency for predefined intents while leveraging RAG for complex or ambiguous queries. The inclusion of a dialogue context manager guarantees coherence across multi-turn interactions, and a feedback-driven mechanism continuously refines intents and confidence thresholds over time. 

Experimental results demonstrated the proposed framework's ability to balance accuracy (95\%), response latency (180ms), and cost efficiency (0.7), while achieving superior context retention and turn efficiency in multi-turn scenarios. The system effectively handles multi-turn dialogues with minor limitations in long conversations exceeding 10 turns. Our contributions include a scalable, adaptive solution for dynamic conversational AI, addressing key industry challenges such as query diversity, evolving knowledge bases, and real-time performance requirements. Future research will focus on enhancing multi-turn context management, conducting ablation studies to isolate module contributions, and exploring real-time learning mechanisms for continuous adaptation. This work advances the state-of-the-art in enterprise conversational AI, providing a robust framework for handling complex, multi-turn interactions efficiently.

\section{Limitations and Future Work}
While our system demonstrates strong performance in enterprise customer support scenarios, it is optimized for English language applications and may require adaptation for multilingual deployments. Expanding to other languages introduces challenges such as acquiring labeled training data and handling linguistic variations, which may increase operational costs and training time. Additionally, our intent classifier is trained on domain-specific datasets, and extending to new domains or industries will necessitate retraining with relevant data, impacting both cost and deployment timelines.

Lastly, integrating real-time learning mechanisms that adapt continuously without periodic retraining is an avenue for future exploration, providing a more seamless and cost-effective method for maintaining system relevance over time. Future work will also include studies to isolate the impact of the dialogue context manager and quantify its contribution to system performance, as well as extending our framework to support multilingual conversations by improving intent recognition and retrieval efficiency across diverse languages.

\bibliography{custom}
\bibliographystyle{acl_natbib}

\appendix

\section{Appendix}
\label{sec:appendix}
\begin{table*}[th!]
\centering
\begin{tabular}{|l|l|l|p{3.5cm}|}
\hline
\textbf{Scenario}        & \textbf{Query Type}                  & \textbf{Response Type}          & \textbf{Impact}                     \\ \hline
Predefined FAQ           & High-confidence intent               & Canned Response                 & Reduced Latency, Cost Savings       \\ \hline
Contextual Query         & Low-confidence intent                & Hybrid (RAG + canned)           & Increased Coherence, Cost Saving     \\ \hline
Out-of-Domain Query      & Undefined intent                     & Full RAG pipeline               & Increased Accuracy                   \\ \hline
\end{tabular}
\caption{Query Handling Scenarios in the Hybrid Framework.}
\label{tab:scenarios}
\end{table*}

\begin{table*}[th!]
\centering
\renewcommand{\arraystretch}{1.5}
\begin{tabular}{|l|l|c|c|c|}
\hline
\textbf{Framework}        & \textbf{Category}      & \makecell{\textbf{Accuracy} \\ (\%)} & \makecell{\textbf{Response Latency} \\ (ms)} & \makecell{\textbf{Cost Efficiency} \\ } \\ \hline
\multirow{3}{*}{Canned Response } & Predefined FAQ       & 93                     &65                             & 1.00                                   \\ \cline{2-5}
                           & Contextual            & 49                     & 65                             & 1.00                                   \\ \cline{2-5}
                           & Out-of-Domain         & 5                     & 75                             & 0.08                                   \\ \hline
\multirow{3}{*}{RAG}       & Predefined FAQ       & 91                     & 376                            & 0.31                                   \\ \cline{2-5}
                           & Contextual            & 92                     & 381                            & 0.31                                   \\ \cline{2-5}
                           & Out-of-Domain         & 90                     & 379                           & 0.31                                   \\ \hline
\multirow{3}{*}{Proposed Framework} & Predefined FAQ       & 96                     & 65                            & 1.00                               \\ \cline{2-5}
                           & Contextual            & 96                     & 182                            & 0.67                                   \\ \cline{2-5}
                           & Out-of-Domain         & 93                     & 379                           & 0.32                                  \\ \hline
\end{tabular}
\caption{Performance comparison of different frameworks across various categories. Baseline cost efficiency is established using average latency and accuracy for canned responses across the entire evaluation dataset as mentioned in Table~\ref{tab:results}.}
\label{tab:framework_comparison}
\end{table*}

\subsection{Extended Workflow}
\label{sec:extended_workflow}

\paragraph{Feedback-Driven Adaptability.} The feedback rates are used to dynamically change the thresholds defined as follows:

\[
\text{NFR} = \frac{\text{NegativeFeedback}}{\text{TotalQueries}}
\]
\[
\text{PFR} = \frac{\text{PositiveFeedback}}{\text{TotalQueries}}
\]

New intents are generated from user feedback and query patterns, which are processed offline to update the Intent Data Store. Intent classification is refined continuously by an adaptive system feedback loop. Unresolved queries are logged, clustered on the basis of semantic similarity, and flagged for review. When a cluster reaches a certain size, a new intent is created offline and integrated into the classifier. Additionally, confidence thresholds are periodically adjusted based on user feedback to improve the routing of ambiguous queries.

\subsubsection{Intent Evolution through Feedback}  
In addition to threshold tuning, the system expands its intent data store based on observed usage patterns and unresolved queries. The intent creation process operates in the following stages:

\begin{enumerate}
    \item \textbf{Logging and Clustering}: All unhandled queries are logged and grouped using semantic similarity clustering.
    \item \textbf{Pattern Detection}: If a cluster of unresolved queries exceeds a predefined frequency threshold, it is flagged for intent creation.
    \item \textbf{New Intent Generation}: A new intent is proposed \& validated by SMEs, and added to the Intent Data Store.
\end{enumerate}

This process ensures that frequently occurring unresolved queries are automatically handled by the intent classifier going forward, thereby improving future query routing.

\textbf{Improving FAQ Classification via Threshold Adjustment}

\textbf{Query:} \textit{“Why am I seeing high costs for my Oracle Autonomous Database instance?”}  

The system classifies this as an FAQ and responds:  \\
\textbf{Response:} \textit{"Oracle Autonomous Database costs depend on the compute shape, storage capacity, and workload type. You can adjust your settings to optimize cost."}  

However, users frequently provide negative feedback, indicating that the response lacks details on Auto Scaling, Always Free tier limits, and OCI pricing policies. This causes NFR to increase, leading to an increase in $\tau_{\text{FAQ}}$. The system becomes more selective in assigning queries to FAQs. More complex cost-related queries are routed to context-aware retrieval mechanisms rather than FAQs.

\textbf{Intent Creation for Repeated OOD Queries}  
\textbf{Query:}  \textit{“How do I configure OCI Object Storage to replicate data to another region?”}  

Initially, the system classifies this as Out-of-Domain (OOD), as no existing intent covers cross-region object storage replication. However, after multiple users ask similar questions, the system clusters these unresolved queries. Once the cluster surpasses the predefined frequency threshold, it is flagged for new intent creation by SMEs:  

\textbf{New Intent:} \textit{“OCI Object Storage Cross-Region Replication”}\\  
\textbf{Associated Response:} \textit{“Detailed steps to enabled and configure cross-region replication as determined by SME”}  

The system proactively resolves similar future queries by classifying them under the newly created intent. Users receive accurate responses immediately instead of being redirected to general support.

\paragraph{Proposed Framework.} The workflow of the proposed system is summarized in Table \ref{tab:scenarios}

\subsection{Detailed Performance Comparison}
\label{appendix:detail_comparison}
This section provides a detailed breakdown of the performance of the proposed hybrid framework compared to baseline systems (Canned Response System) and RAG Pipeline across different query categories: Predefined FAQs, Contextual Queries, and Out-of-Domain Queries. The metrics include accuracy, response latency, and cost efficiency, highlighting the strengths and trade-offs of each approach.

\paragraph{Analysis} The results in Table \ref{tab:framework_comparison} demonstrate the trade-offs between accuracy, latency, and cost efficiency:
\begin{itemize}
    \item \textbf{Predefined FAQs:} The proposed framework achieves a balance, with similar accuracy with the canned-response system while reducing latency by ~82\% compared to the RAG pipeline.
    \item \textbf{Contextual Queries:} The proposed framework strikes a balance between RAG's accuracy (92\%) and the canned-response system's latency (65ms), achieving 96\% accuracy with an acceptable latency of 182ms.
    \item \textbf{Out-of-Domain Queries:} The RAG Pipeline and the proposed framework have a very similar latency and performance with our proposed framework have slight better accuracy (3\%) owing to the better handling of context and queries.
    % While the RAG pipeline leads in accuracy, the hybrid framework offers a latency reduction of 60\% with only a slight drop in accuracy (78\% vs. 75\%).
\end{itemize}

\begin{table*}[th!]
\centering
\renewcommand{\arraystretch}{1.2}
\begin{tabular}{|l|l|c|}
\hline
\textbf{Query}                               & \textbf{Category}     & \textbf{Confidence Level} \\ \hline
How do I reset my password?                  & Predefined FAQ         & 0.95                      \\ \hline
What are the steps to integrate autoscaling? & Contextual             & 0.70                      \\ \hline
What are compliance requirements for data?   & Out-of-Domain          & 0.40                      \\ \hline
Can you elaborate on scaling options?        & Multi-Turn Follow-Up   & 0.75                      \\ \hline
\end{tabular}
\caption{Sample Queries from the in-house Dataset.}
\label{tab:synthetic_queries}
\end{table*}

\begin{table*}[th!]
\centering
\renewcommand{\arraystretch}{1.2}
\begin{tabular}{|l|p{5cm}|l|p{6cm}|}
\hline
\textbf{Turn} & \textbf{User Query}                                  & \textbf{Framework}      & \textbf{System Response}                                      \\ \hline
1             & What are the steps to enable advanced analytics?    & Canned Response         & Analytics can be enabled in the dashboard settings.           \\ \hline
2             & Can you explain what metrics are available?          & Hybrid Response         & Available metrics include user engagement, retention, and revenue. \\ \hline
3             & How can I visualize these metrics effectively?       & RAG Response            & Visualization tools like Tableau and Power BI integrate seamlessly with our platform. \\ \hline
4             & What steps are required to connect Tableau?          & Hybrid Response         & Refer to the integration settings under "Analytics" and provide your Tableau API key. \\ \hline
5             & Are there any tutorials for advanced analytics setup?& RAG Response            & Yes, detailed tutorials can be found in the documentation section under "Advanced Analytics." \\ \hline
\end{tabular}
\caption{Multi-Turn Example Showcasing Evolving Intents and Follow-Ups.}
\label{tab:multi_turn_examples}
\vspace{-1em}
\end{table*}

\subsection{In-House Dataset Overview}
\label{sec:in-house-data}
The evaluation leveraged a in-house dataset on customer support for OCI Cloud based Services of 10,000 queries across three categories: predefined FAQs, contextual queries, and out-of-domain queries. Table~\ref{tab:synthetic_queries} provides a sample of the queries used in the evaluation.

For BERT fine-tuning, we used in-house conversational dataset which is domain specific, with 35,000 human-customer conversations curated over a period of 6 months.

\subsection{Multi-Turn Interaction Examples}
To demonstrate the framework’s adaptability, Table~\ref{tab:multi_turn_examples} outlines examples of evolving user queries and how the system dynamically adapts to maintain coherence.

\subsection{Failure Cases and Error Analysis}
\label{appendix:Failure_Cases}
We conducted a manual error analysis on 500 dialogue samples spanning diverse user intents. Three independent annotators with experience in enterprise conversational AI systems evaluated these dialogues, with an inter-annotator agreement of 0.91 (Cohen’s Kappa). Inter-annotator agreement was calculated by comparing the categorical labels assigned (out of 4 shown below) by each annotator across all 500 dialogue samples. Annotators independently labeled each dialogue, and disagreements were resolved through discussion to refine the labeling criteria. The high agreement score (0.91) reflects consistency in identifying and categorizing errors across evaluators.

Errors were categorized as follows:
\begin{itemize}
    \item \textbf{Edge Cases in Intent Classification (21\% of errors):} Queries were misclassified due to subtle semantic differences. For example, the query \textit{“Can you assist with integrating API X for multi-platform deployment?”} was routed to a general FAQ response about API usage due to high lexical similarity.
    \item \textbf{Long Multi-Turn Dialogues (35\% of errors):} In conversations exceeding 10 turns, latency increased, and context tracking sometimes failed. For instance, after handling a billing query, the system mistakenly retained billing context when the user shifted to technical support.
    \item \textbf{Retrieval Inaccuracy (25\% of errors):} Some queries led to incomplete or off-topic document retrieval. For example, a query like \textit{“Provide the latest number of regions your cloud service is available in”} retrieved outdated documents due to incomplete index updates.
    \item \textbf{Feedback Misalignment (19\% of errors):} User feedback was sometimes misinterpreted. For instance, a user rated a correct response poorly due to slow response time rather than content accuracy, leading to unnecessary adjustments in the intent classifier.
\end{itemize}

Table \ref{tab:failure_cases} summarizes these failure cases and suggested remedies. This detailed error analysis highlights both the strengths of our system and areas for future improvement.

\begin{table*}[h!]
\centering
\renewcommand{\arraystretch}{1.8} % Increase row height
\scalebox{0.8}{
\begin{tabular}{|l|p{4.5cm}|p{4.5cm}|p{2.5cm}|} % Adjusted column widths
\hline
\textbf{Scenario}            & \textbf{Issue}                                                     & \textbf{Remedy}                                                   & \textbf{Error Distribution (\%)} \\ \hline
Edge Cases in Intent Classification       & Query incorrectly routed to canned responses                     & A stronger fallback strategy could improve reliability                      & 21\% (7/32) \\ \hline
Long Multi-Turn Dialogues         & Latency for very long conversations       & Optimize dialog context manager to reduce latency.             & 35\% (11/32) \\ \hline
Retrieval Inaccuracy & Incomplete or outdated documents retrieved & Regular index updates and improved retrieval ranking & 25\% (8/32) \\ \hline
Feedback Misalignment & User feedback misinterpreted during adjustments & Implement context-aware feedback processing & 19\% (6/32) \\ \hline
\end{tabular}}
\caption{Failure Cases and Suggested Remedies. A total of 32 erroneous dialogues were identified out of 500 tested samples.}
\label{tab:failure_cases}
\end{table*}

\subsection{Prototype Implementation}
The framework is implemented as a modular system using microservices:
\begin{itemize}
    \item \textbf{Frontend:} Built with React.js for user interaction and feedback collection.
    \item \textbf{Backend:} Flask microservices handle query classification, retrieval, and feedback processing~\cite{grinberg2018flask}.
    \item \textbf{Storage:} Elasticsearch stores canned responses \& external knowledge base, while FAISS and dense embeddings support retrieval~\cite{karpukhin2020dense}. 
    \item \textbf{Memory Cache:} A memory-augmented module maintains embeddings of prior inter-actions in OCI Cache (Managed Redis), allowing the system to retain relevant historical context across multiple dialogue turns.
    \item \textbf{Model Deployment:} Models (e.g., BERT, Encoder, Cross-Encoder, GPT-3 \& other proprietary LLMs) are deployed using in-house architecture and OCI Gen AI Service for scalability.
\end{itemize}

\section{Technical Implementation of Multi-Turn Adaptation}
\label{sec:multi_turn_appendix}
The \textbf{Dialogue Context Manager} is implemented using a transformer-based architecture with the following components:
\begin{itemize}
    \item \textbf{Context Embeddings:} Queries are encoded using fine-tuned BERT embeddings capturing semantic information and historical contexts are encoded using an in-house Bi-Encoder.
    \item \textbf{Memory Module:} A memory-augmented module maintains embeddings of prior interactions in cache (Redis), allowing the system to retain relevant historical context across multiple dialogue turns.
    \item \textbf{Context Attention Mechanism:} An attention layer prioritizes recent or semantically relevant interactions, dynamically retrieving context embeddings as input to the intent classifier and response generator.
    \item \textbf{Sliding Context Window:} Implements a fixed-length sliding window to limit the memory footprint and computational complexity by retaining only the most relevant context from prior turns.
\end{itemize}

The context manager utilizes the embeddings and attention scores to generate a composite representation of the current dialogue state, which is passed to downstream components, such as the hybrid response generator. The dynamic adaptation ensures responses remain coherent and contextually grounded in multi-turn settings.

\section{Technical Implementation of Hybrid Routing}
\label{sec:routing_appendix}
Hybrid routing combines canned responses and RAG outputs using a confidence-based decision-making pipeline:
\begin{itemize}
    \item \textbf{Confidence Scoring:} The intent classifier assigns a confidence score to each query based on the similarity between the query embedding and predefined intent embeddings.
    \item \textbf{Thresholding Mechanism:} Queries with a confidence score above a pre-defined threshold (e.g., 85\%) are routed to the canned response repository for rapid resolution.
    \item \textbf{Response Generation:} For low-confidence queries or multi-turn scenarios, responses are generated by blending canned responses with retrieved content from the RAG pipeline. Sample prompt used for blending the responses using confidence scores is shown in Figure \ref{fig:prompt}
    \begin{figure*}[t!]
    \centering
    \includegraphics[width=\textwidth]{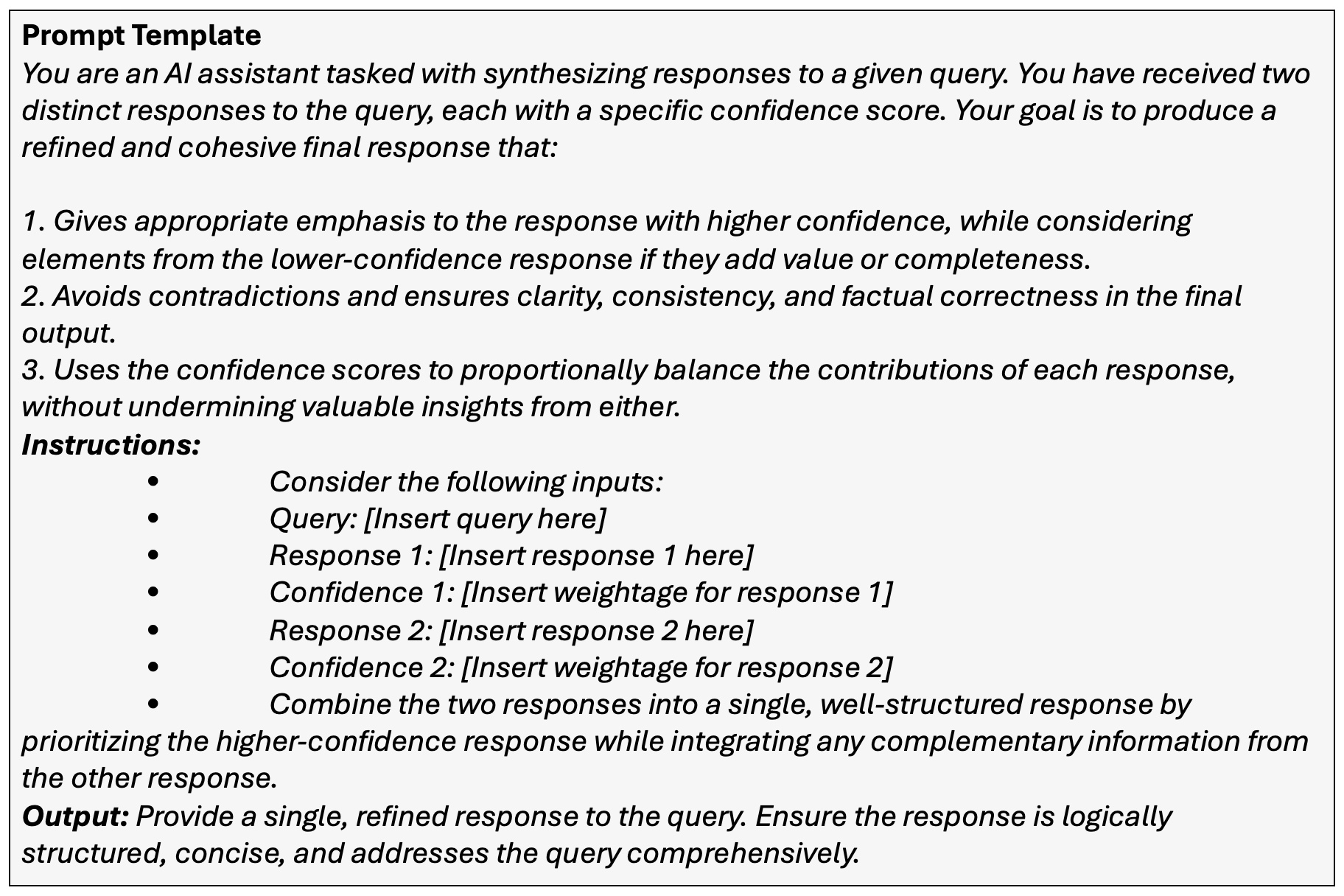} % Placeholder for system architecture image
    \caption{Prompt for Blending Responses}
    \label{fig:prompt}
    \vspace{-1em}
\end{figure*}

\end{itemize}
This mechanism optimizes query handling for diverse scenarios while ensuring minimal latency and maximal accuracy.

\section{Feature Limitations \& Related Future Work}
\subsection{Limitations}
Despite the strong performance of the proposed framework on a variety of metrics, certain feature-specific limitations remain:
\begin{itemize}
    \item \textit{Edge Cases in Intent Classification}: Ambiguous queries near confidence thresholds may cause inconsistencies, as seen in our error analysis, where subtle semantic differences led to misclassification. A stronger fallback strategy could improve reliability.
    \item \textit{Latency in Long Multi-Turn Dialogues}: Latency issues for very long conversations (over 10 turns) were identified in 30\% of errors, highlighting the need to optimize the dialogue context manager for faster context updates.
    \item \textit{Retrieval Inaccuracy}: Incomplete or outdated document retrieval (20\% of errors) due to index inconsistencies highlights the need for regular index updates and improved retrieval ranking.
    \item \textit{Feedback Misalignment}: User feedback misinterpretation (10\% of errors) occasionally led to suboptimal adjustments, suggesting the need for context-aware feedback processing.
\end{itemize}

\subsection{Future Work}
Future research could address these limitations by:
\begin{itemize}
    \item Developing advanced intent detection techniques and fallback mechanisms to handle ambiguous and low-confidence queries more effectively.
    \item Enhancing multi-turn context tracking with memory-augmented models to improve coherence across long dialogues.
    \item Implementing regular index updates and fine-tuned retrieval processes to ensure accurate document retrieval.
    \item Integrating context-aware feedback processing to ensure accurate adaptation of system responses based on user ratings.
    \item Exploring distributed architectures and load-balancing techniques for scalability under peak query loads.
\end{itemize}

\end{document}